\documentclass[12pt]{article}

\usepackage[
  margin=1.5cm,
  includefoot,
  footskip=30pt,
]{geometry}
\counterwithin{figure}{section}
\usepackage[table]{xcolor}
\usepackage{tikz}
\usetikzlibrary{matrix, arrows}
\usepackage{amsmath,amssymb}
\usepackage{amsthm}
\usepackage{mathtools}
\usepackage{xspace}
\usepackage[noend]{algorithmic}
\usepackage[ruled,vlined]{algorithm2e}
\usepackage{url}
\usepackage{makeidx}
\usepackage{enumerate}
\usepackage{epstopdf}
\usepackage{booktabs}
\usepackage{color}
\usepackage[utf8]{inputenc}
\usepackage{thm-restate}
\usepackage{scalerel,stackengine}
\usepackage[shortlabels]{enumitem}
\usepackage{xr}
\usepackage{fancyvrb}
\usepackage{bold-extra}
\usepackage[width=474.18663pt]{caption}
\usepackage{subfigure}
\usepackage[most]{tcolorbox}
\usepackage{fvextra}
\usepackage{float}
\usepackage{alltt}
\usepackage{soul}
\usepackage{fancyvrb}
\usepackage{multirow}
\usepackage{array}
\usepackage{threeparttable}
\usepackage{authblk}
\usepackage{multicol}
\usepackage{xcolor}
\usepackage{colortbl}
\usepackage{siunitx}
\usepackage{fancyvrb}
\usepackage[final]{hyperref}
\usepackage[bottom]{footmisc}

\newcommand{\DV}{GPT-4\xspace}
\usepackage{listings}
\usepackage{tikz}
\usetikzlibrary{shapes,calc,positioning}

\global

\tcbset{
  aibox/.style={
    width=474.18663pt,
    top=10pt,
    colback=white,
    colframe=black,
    colbacktitle=black,
    enhanced,
    center,
    attach boxed title to top left={yshift=-0.1in,xshift=0.15in},
    boxed title style={boxrule=0pt,colframe=white,},
  }
}
\newtcolorbox{AIbox}[2][]{aibox,title=#2,#1}

\definecolor{Gray}{gray}{0.95}
\definecolor{aigold}{RGB}{244,210, 1} 
\definecolor{aigreen}{RGB}{210,244,211} 

\sethlcolor{aigreen}

\definecolor{aired}{RGB}{255,180,181}

\newtcbox{\mybox}[1][green]{on line,
arc=0pt,outer arc=0pt,colback=#1!10!white,colframe=#1!50!black,
boxsep=0pt,left=0pt,right=0pt,top=0pt,bottom=0pt,
boxrule=0pt,bottomrule=0pt,toprule=0pt}

\title{Capabilities of {\DV} on Medical Challenge Problems}
\author[1]{Harsha Nori}
\author[1]{Nicholas King}
\author[2]{Scott Mayer McKinney}
\author[1]{\\Dean Carignan}
\author[1]{Eric Horvitz}
\date{}
\affil[1]{Microsoft}
\affil[2]{OpenAI}
\begin{document}

\maketitle

\begin{abstract}
    Large language models (LLMs) have demonstrated remarkable capabilities in natural language understanding and generation across various domains, including medicine. We present a comprehensive evaluation of GPT-4 \cite{openai2023gpt4}, a state-of-the-art LLM, on medical competency examinations and benchmark datasets. GPT-4 is a general-purpose model that is not specialized for medical problems through training or engineered to solve clinical tasks. Our analysis covers two sets of official practice materials for the United States Medical Licensing Examination (USMLE), a three-step examination program used to assess clinical competency and grant licensure in the United States. We also evaluate performance on the MultiMedQA suite of benchmark datasets. Beyond measuring model performance, experiments were conducted to investigate the influence of test questions containing both text and images on model performance, probe for memorization of content during training, and study calibration of the probabilities, which is of critical importance in high-stakes applications like medicine. Our results show that GPT-4, without any specialized prompt crafting, exceeds the passing score on USMLE by over 20 points and outperforms earlier general-purpose models (GPT-3.5) as well as models specifically fine-tuned on medical knowledge (Med-PaLM, a prompt-tuned version of Flan-PaLM 540B). In addition, GPT-4 is significantly better calibrated than GPT-3.5, demonstrating a much-improved ability to predict the likelihood that its answers are correct. We also explore the behavior of the model qualitatively by presenting a case study that shows the ability of GPT-4 to explain medical reasoning, personalize explanations to students, and interactively craft new counterfactual scenarios around a medical case. Implications of the findings are discussed for potential uses of GPT-4 in medical education, assessment, and clinical practice, with appropriate attention to challenges of accuracy and safety.
\end{abstract}

\section{Introduction}

Large language models (LLMs) have exhibited a remarkable ability to interpret and generate sequences across a wide array of domains, such as natural language, computer code, and protein sequences. Numerous powerful models are based on the transformer architecture \cite{vaswani2017attend}, adapted to language and trained in a  self-supervised manner \cite{radford2018,devlin2018bert}. Scores on a variety of benchmarks have generally improved with scale, involving increasing model size, dataset size, and the amount of training computation in tandem \cite{kaplan2020scalelaw,HELM2022}. The empirical findings resonate with a theoretical analysis \cite{bubeck2021scaling} which shows the necessity of scale for robustness of inferences from large neural models \cite{bubeck2021scaling}. 

 Over the last several years, LLMs trained on massive, cross-disciplinary corpora have become potent building blocks in the creation of task-focused systems \cite{Bommasani2021}. Methods for refining the models toward a particular domain include fine-tuning with specialized datasets drawn from target applications and general methods for steering the behavior of the models, such as reinforcement learning with human feedback (RLHF), which guides the system toward a better understanding of end-users' requests \cite{christiano2017deep,bai_training_2022}.
 
 There has also been great interest in the ability of LLMs to make useful inferences for a broad range of specialized tasks without dedicated fine-tuning. The performance of general-purpose LLMs using few- or even zero-shot prompting highlights their potential for assisting with tasks across problem types, specialty areas, and disciplines \cite{brown2020language}. Recently, researchers have investigated benchmarks that provide insight into how LLMs encode clinical knowledge and might be harnessed to augment the practice of medicine. Here we compare the performance of the recently released (text-only) {\DV} model with its predecessors in the GPT family on medical challenge problems. While details on measures of scale for {\DV}, including the number of model parameters and the size and scope of training data, have not been made public, it has been reported that both dimensions are significantly bigger than for GPT-3.5, the model behind ChatGPT \cite{openai2023gpt4}. 

Exploration of the capabilities of LLMs on medical problem solving is part of a long-standing research program on AI in medicine, going back to the classic work of Ledley and Lusted \cite{ledley59reasoning}.  Over the decades since, explorations of computational methods for assisting physicians have been marked by shifting enthusiasm for different representations and reasoning methods, including core probabilistic and decision-theoretic methods (e.g., \cite{gorry1968experience,Heckerman1992TowardNE}), rule-based production systems (e.g., \cite{shortliffe1977mycin,buchanan1984rule}), semantic graphs (e.g., \cite{patil1981causal}), supervised learning from databases of medical information (e.g., \cite{wiens2016patient, henry2015targeted, escobar2020automated, caruana2015intelligible}), and deep neural network models (e.g., \cite{esteva2017dermatologist,suresh2017clinical, rajpurkar2017chexnet,mckinney2020}). While the flurry of efforts to use deep learning to attain human-level performance on medical tasks began in the field of computer vision for diagnostics, it has since grown to encompass benchmarks for more general clinical reasoning mediated through natural language. The models deployed in this context may be trained on specific medical corpora or foundation models trained on massive amounts of general language and/or visual information and then adapted to medical data through dedicated fine-tuning.

Our key contribution is to investigate the capabilities of {\DV} on medical challenge problems. To establish strong baselines for comparison, we evaluate {\DV} against GPT-3.5 and reported results from Flan-PaLM 540B.
Our goal is to establish ``out-of-the-box" performance numbers for {\DV}. To that end, we use the simplest prompts possible (a zero-shot and randomly selected 5-shot prompt with direct inference of the answer) and find that {\DV} obtains best-in-class performance without any need for elaborate prompting techniques or domain-specific fine-tuning.

We begin by interrogating the performance of the models on challenge problems developed to assess competencies of medical students and residents. This exploration consists of a comprehensive evaluation of the performance of {\DV} on Steps 1-3 of the United States Medical Licensing Examination (USMLE). The exam is part of the official accreditation protocol though which medical licensure is determined in the U.S. Our results are based on sample exams and self-assessment materials officially published by the National Board of Medical Examiners (NBME). The findings show that zero-shot {\DV} significantly outperforms earlier models, achieving an average score of 86.65\% and 86.7\% on the Self-Assessment and Sample Exam of the USMLE tests, respectively, compared to 53.61\% and 58.78\% for GPT-3.5. After reviewing results for the USMLE studies, we examine several other medical benchmarks. Zero-shot {\DV} significantly outperforms GPT-3.5 and the recently introduced Flan-PaLM 540B model on MultiMedQA\cite{singhal2022large}, a suite of commonly used benchmark datasets in the literature on machine learning for medicine.

Beyond characterizing overall performance, our investigation covers several other facets of LLM behavior in the medical domain. We study the performance of the text-only {\DV} on examination questions that are text-centric versus questions that rely on images. Given that reliable information about the probability of correctness is critical in healthcare and other high-stakes applications, we evaluate the calibration of the probabilities implicitly assigned to answers. We assess evidence that the model has been exposed to (and memorized) the content of the examinations through its training data. We further explore qualitative behavior of the model via a case study that demonstrates the capabilities of {\DV} to explain medical reasoning and interactively support students on counterfactual scenarios around a medical case. Finally, we examine the implications of our findings, including the potential for {\DV} and its successors to help with medical education and to provide assistance to healthcare professionals, taking into consideration concerns related to accuracy, fairness, and broader impacts on the practice of medicine. We particularly reflect on the limitations of benchmark-based performance evaluations, and discuss the precautions and advances needed to make use of models like {\DV} in real world settings. Significant work remains to evaluate these systems comprehensively, and much caution is needed. However, we expect multiple real world uses, such as lower stakes applications that include expert oversight as part of generations and workflows. In the longer-term, we see significant potential for {\DV} and its successors to have a positive impact in medicine.

\section{Methodology}

While {\DV} supports multi-modal capabilities \cite{openai2023gpt4}, our methodology focuses on a text-only version of the model, referred to as {\DV} (no vision) by OpenAI. For simplicity, all subsequent references to {\DV} in this paper refer to the text-only model without vision capabilities. Discussion of how the text-only model performs on questions with images can be found in Section \ref{sec:media-questions}.

\subsection{Datasets}

To evaluate {\DV}, we consider six datasets that cover different aspects of medical knowledge and reasoning. Two of these datasets, the USMLE Sample Exam and USMLE Self Assessments, are sourced directly from the the National Board of Medical Examiners (NBME), the organization that governs and administers the the examination process. The other four datasets, MedQA, PubMedQA, MedMCQA, and MMLU, are publicly available benchmarks that contain questions based on medical literature, clinical cases, and user-generated content. These four datasets have been widely used to benchmark LLM performance on medical reasoning tasks. All four datasets constitute a large part of the recently introduced ``MultiMedQA" benchmark \cite{singhal2022large}. Details on each dataset are provided in Appendix \ref{sec:dataset-descriptions}.

\subsection{Prompting}
\begin{figure}[H]
\begin{AIbox}{Prompt template for multiple choice questions}
The following are multiple choice questions (with answers) about medical knowledge. \\
\verb|{{few_shot_examples}}|

\verb|{{context}}|**Question:** \verb|{{question}}|
\verb|{{answer_choices}}|
**Answer:**( 
\end{AIbox}
\caption{Template used to generate prompts on all multiple choice questions (from \cite{singhal2022large}). Elements in double braces \{\{\}\} are replaced with question-specific values.}
\label{fig:prompt-template}
\end{figure}

\begin{figure}[H]
\begin{AIbox}{Sample question using prompt template}
The following are multiple choice questions (with answers) about medical knowledge. \\
**Question**: A 40-year-old woman has had hypercalcemia for 1 year and recently passed a renal calculus. Serum parathyroid hormone and calcium concentrations are increased, and serum phosphate concentration is decreased. Parathyroid hormone most likely causes an increase in the serum calcium concentration by which of the following mechanisms? \\
(A) Decreased degradation of 25-hydroxycholecalciferol \\
(B) Direct action on intestine to increase calcium absorption \\
(C) Direct action on intestine to increase magnesium absorption \\
(D) Increased synthesis of 25-hydroxycholecalciferol \\
(E) Inhibition of calcitonin production \\
(F) Stimulation of 1,25-dihydroxycholecalciferol production \\
**Answer:**(\hl{F}
\end{AIbox}
\caption{Instantiated example of Figure \ref{fig:prompt-template}. {\DV}'s (correct) response is shown in \hl{green}.}
\label{fig:prompt-example}
\end{figure}

To establish baseline model performance, and provide fair comparisons, we employ the exact same prompting structure as \cite{singhal2022large}. An example of the template and a fully instantiated prompt are shown in Figures \ref{fig:prompt-template} and \ref{fig:prompt-example}, respectively. In the zero-shot setting, the \verb|few_shot_examples| slot is simply left blank. Similarly, for datasets that don't provide additional context for each question, we leave the \verb|context| slot blank as well. For models optimized for chat based scenarios, like ChatGPT and {\DV}, we make minor modifications to this template to simulate a conversation. Examples of the few-shot and chat-based versions of the prompts are presented in Appendix \ref{sec:app-prompts}.

Our prompting structure enables us to benchmark more efficiently by using minimal context tokens and a single-generation token for each inference. Furthermore, we take advantage of the \verb|logit_bias| parameter in OpenAI's API to induce the model to generate only valid responses. For example, on a question with 4 multiple choice answers, A, B, C, and D, we pass:

\begin{center}
\begin{BVerbatim}
logit_bias = {32 : 25, 33 : 25, 34 : 25, 35 : 25}
openai.completion.create(...logit_bias = logit_bias, ...)
\end{BVerbatim}
\end{center}

where 32-35 are the tokens that correspond to the letters A-D, respectively. 

Our goal is to measure the baseline performance of {\DV} on medical multiple-choice questions (MCQs) using a simple approach, without resorting to complex methods such as chain-of-thought prompting \cite{wei2022chain}, retrieval augmented generation \cite{nakano2021webgpt}, or ensembling strategies \cite{wang2022self}. In prior work, these methods have shown to enhance the performance of LLMs on medical MCQs considerably\cite{singhal2022large, wei2022chain}. However, we show that {\DV} can attain outstanding results even without these techniques, exceeding both human performance levels and that of other models employing sophisticated prompting methods. We leave exploration of the larger space of performance optimization to future work.

\subsection{Model Comparison}

We evaluate both {\DV} and its predecessor model, GPT-3.5, against all benchmark datasets studied in this paper. For each model, we consider both a zero-shot and 5-shot prompt following the template described in Figure \ref{fig:prompt-template}. For zero-shot evaluations, we directly report accuracy on each dataset. In the 5-shot setting, we report leave-one-out cross validation (LOOCV) \cite{hastie2009elements} accuracy, where for each evaluation sample, we draw the 5 few-shot exemplars randomly from the remainder of the dataset. We also incorporate results reported in the literature from other models evaluated on the same datasets, including ChatGPT, InstructGPT, Flan-PaLM 540B, and Med-PaLM. We note that Flan-PaLM 540B and Med-PaLM are currently unavailable for public consumption; therefore, any results reported on these models are sourced directly from \cite{singhal2022large}.

\section{Performance of {\DV} on Medical Competency Exams}

We analyze model performance on two sets of official practice materials for the United States Medical Licensing Examination (USMLE). USMLE is a three-step examination program employed to assess clinical competency.  Licensure for independent provision of healthcare in the United States requires passing the exam sequence. Each step marks a milestone in medical training. Step 1 of the USMLE examination is typically taken by medical students after completing their preclinical training. Step 1 covers core clinical knowledge, including pathology and physiology, and the basis for medical conditions.  Step 2, taken at the completion of an MD program, tests clinical understanding by probing the test-takers' knowledge about diagnosis and patient management.  Scores on Step 2 are often considered in decisions about interviews and acceptance into residency programs.  Step 3 is the final examination in the USMLE sequence. The exam assesses medical residents' ability to apply their working knowledge of medicine in the unsupervised practice of medicine. The test probes working clinical and biomedical knowledge deemed as essential for taking on independent responsibility for providing general medical care. Passing performance on Step 3 is required for being licensed to practice medicine without supervision.

We emphasize that the materials we used in this section of our benchmarks are officially purchased and sourced from the National Board of Medical Examiners (NBME), which is one of the organizations that develops and administers the USMLE. While some publicly available datasets and papers rely on unofficial, third-party sources to approximate USMLE performance (e.g., the MedQA dataset\footnote{See the discussion about MedQA and other unofficial datasets from the USMLE: \href{https://www.usmle.org/usmle-program-discusses-chatgpt}{https://www.usmle.org/usmle-program-discusses-chatgpt}}, evaluated in section \ref{sec:med-benchmarks}), we believe the USMLE Sample Exam and USMLE Self Assessment are among the most authentic sources of exam questions available for this type of study. More details about the datasets can be found in Appendix \ref{sec:dataset-descriptions}, with concerns about model memorization discussed in Section \ref{sec:memorization}.

\subsection{Results}

{\DV} shows a remarkable improvement over its predecessor models on official USMLE exam questions, improving by over 30 percentage points on both exams when compared to GPT-3.5. We also find that {\DV} shows a similarly precipitous improvement over recent independently reported performance metrics for ChatGPT, a popular variant of GPT-3.5 optimized for chat-based interaction \cite{kung2023performance}. Med-PaLM and Flan-PaLM 540B are not currently available for public use, so we are unable to report their performance on these particular datasets. Comparisons against previously reported results from the PaLM family of models \cite{singhal2022large} are available in Section \ref{sec:med-benchmarks}. 

The USMLE website states that, while specific pass thresholds vary every year, examinees must answer approximately 60 percent of multiple-choice questions correctly to achieve a passing score\footnote{\href{https://www.usmle.org/scores-transcripts}{https://www.usmle.org/scores-transcripts}}.
While earlier models like GPT-3.5 were approaching the passing threshold, {\DV} clears this bar by a large margin. 


\begin{table}[H]
\centering
\label{tab:self-assessment-scores}
\caption{Comparison of performance of models on the USMLE Self Assessment. {\DV} significantly outperforms GPT-3.5.}

\begin{threeparttable}
    \begin{tabular}{l>{\bfseries}cc|cc}
        \toprule 
        \multirow{2}{*}{\parbox{4cm}{\centering USMLE\\Self Assessment}} & \multicolumn{1}{c}{\textbf{{\DV}}} & \multicolumn{1}{c}{{\DV}} & \multicolumn{1}{c}{GPT-3.5} & \multicolumn{1}{c}{GPT-3.5} \\ 
        & \multicolumn{1}{c}{\textbf{(5 shot)}} & \multicolumn{1}{c}{(zero shot)} & \multicolumn{1}{c}{(5 shot)} & \multicolumn{1}{c}{(zero shot)} \\
        \midrule 
        Step 1 & 85.21 & 83.46 & 54.22 & 49.62 \\ 
        Step 2 & 89.50 & 84.75 & 52.75 & 48.12 \\
        Step 3 & 83.52 & 81.25 & 53.41 & 50.00 \\
        \midrule 
        Overall Average\tnote{*} & 86.65 & 83.76 & 53.61 & 49.10 \\
        \bottomrule 
    \end{tabular}
    \begin{tablenotes}
        \item[*] Calculated as $\frac{\# correct}{\# questions}$ across all three steps. Each step has slightly different sample size.
    \end{tablenotes}
\end{threeparttable}
\end{table}


\begin{table}[H]
\centering
\label{tab:sample-exam-scores} 
\caption{Comparison of performance of models on the USMLE Sample Exam. This dataset is considered in \cite{kung2023performance}. {\DV} significantly outperforms both GPT-3.5 and independently reported ChatGPT scores.}

\begin{threeparttable}
    \begin{tabular}{l>{\bfseries}cc|cc|c}
        \toprule 
        \multirow{2}{*}{\parbox{4cm}{\centering USMLE\\Sample Exam}} & \multicolumn{1}{c}{\textbf{{\DV}}} & \multicolumn{1}{c}{{\DV}} & \multicolumn{1}{c}{GPT-3.5} & \multicolumn{1}{c}{GPT-3.5} & \multicolumn{1}{c}{ChatGPT\tnote{†}} \\ 
        & \multicolumn{1}{c}{\textbf{(5 shot)}} & \multicolumn{1}{c}{(zero shot)} & \multicolumn{1}{c}{(5 shot)} & \multicolumn{1}{c}{(zero shot)} & \multicolumn{1}{c}{(zero shot)} \\
        \midrule 
        Step 1 & 85.71 & 80.67 & 52.10 & 51.26 & 55.1 \\ 
        Step 2 & 83.33 & 81.67 & 58.33 & 60.83 & 59.1 \\
        Step 3 & 90.71 & 89.78 & 64.96 & 58.39 & 60.9 \\
        \midrule 
        Overall Average\tnote{*} & 86.70 & 84.31 & 58.78 & 56.91 & --\\
        \bottomrule 
    \end{tabular}
    \begin{tablenotes}
        \item[*] Calculated as $\frac{\# correct}{\# questions}$ across all three steps. Each step has slightly different sample size.
        \item[†] ChatGPT-3.5 as tested by \cite{kung2023performance}. We report scores from their "MC-NJ encoding with indeterminate responses censored" setting as the most similar to ours. Note that \cite{kung2023performance} removes questions with media elements while we do not.
    \end{tablenotes}
\end{threeparttable}

\end{table}

\subsubsection{Language- and Vision-Centric Challenges}
\label{sec:media-questions}

The performance of the {\DV} model (without vision) on the USMLE Self Assessment and Sample Exam is especially surprising as both exams make frequent use of media elements (e.g. graphs, photographs, charts) in their questions, which do not get passed to the model. In a manual labeling exercise, we found that the Self Assessment had 314 questions with references to media out of 2173 total (14.4\% of the dataset), while the Sample Exam had 49 questions referencing media out of 376 total (13.0\% of the dataset). 

\begin{table}[H]
\centering
\caption{Accuracy on questions that reference visual media (images and graphs) that were not passed to the model, compared to accuracy on questions with pure text.} 
\begin{threeparttable}
        \begin{tabular}{ll>{\bfseries}cccc}
        \toprule 
        \multirow{2}{*}{{\parbox{4cm}{\centering Dataset}}} & \multirow{2}{*}{Question Type} & \multicolumn{1}{c}{\textbf{{\DV}}} & \multicolumn{1}{c}{{\DV}} & \multicolumn{1}{c}{GPT-3.5} & \multicolumn{1}{c}{GPT-3.5} \\ 
        & & \multicolumn{1}{c}{\textbf{(5 shot)}} & \multicolumn{1}{c}{(zero shot)} & \multicolumn{1}{c}{(5 shot)} & \multicolumn{1}{c}{(zero shot)} \\ 
        \midrule 
        \multirow{3}{*}{\parbox{4cm}{\centering USMLE\\Self Assessment}} & Text & 89.51 & 86.39 & 55.30 & 50.40\\
        & Media & 69.75 & 68.15 & 43.63 & 41.40 \\
        & All & 86.65 & 83.76 & 53.61 & 49.10 \\
        \midrule
        \multirow{3}{*}{\parbox{4cm}{\centering USMLE\\Sample Exam}}    & Text & 87.77 & 85.63 & 59.63 & 57.80\\
        & Media & 79.59 & 75.51 & 53.06 & 51.02 \\
        & All & 86.70 & 84.31 & 58.78 & 56.91  \\
        \bottomrule
    \end{tabular}
\end{threeparttable}

\label{tab:vision-sample-exam} 
\end{table}

A segmented performance analysis of these questions is presented in Table \ref{tab:vision-sample-exam}. While {\DV} (no vision) performs best on questions that contain only text, it still performs well on questions with media elements, obtaining 70-80\% prediction accuracies for these questions on both exams. 

Investigating this phenomena further revealed that, despite not being able to see relevant images, the text-only {\DV} model is often able to employ logical reasoning and test-taking strategies to select a reasonable answer choice. A demonstrative example of our interactive investigation is shown in Figure \ref{fig:media-interactive}.

\begin{figure}[H]
\begin{AIbox}{{\DV}: Solving a question without seeing media.}
{\scriptsize
QUESTION: A 45-year-old man comes to the physician because of a 10-year history of heartburn that occurs after he eats late at night or consumes spicy food. He has had no weight loss or difficulty swallowing. He takes over-the-counter antacids as needed, but they relieve his discomfort only temporarily. Physical examination shows no abnormalities. An endoscopy is done. The distal esophagus is shown in the photograph. Which of the following is the most likely cause of this patient's symptoms? 

ANSWER CHOICES: 
``A": ``Defect in secretin production" 
``B": ``Excessive gastrin production" 
``C": ``Excessive transient lower esophageal relaxations" 
``D": ``Failure of primary esophageal peristalsis" 
``E": ``Failure of saliva production" 

ANSWER: \hl{C} \\
----- \\
Q: ``You are a large language model that just answered the following question. Why did you pick the answer choice C?”

A: ``\hl{The answer choice C, ``Excessive transient lower esophageal relaxations," is the most likely cause of this patient's symptoms because it is the most common cause of gastroesophageal reflux disease (GERD). GERD is a condition in which stomach acid flows back into the esophagus, causing heartburn and other symptoms. The lower esophageal sphincter (LES) is a ring of muscle at the bottom of the esophagus that normally prevents stomach acid from flowing back into the esophagus. Transient lower esophageal relaxations (TLESRs) are episodes in which the LES relaxes inappropriately, allowing stomach acid to flow back into the esophagus. TLESRs are the most common cause of GERD, and they are often triggered by factors such as eating late at night or consuming spicy food, as described in the patient's history. The other answer choices are less likely to be the cause of this patient's symptoms."} \\
----- \\
Q: ``Doesn't the question mention a photograph? How did you come up with the answer without seeing the photo?”
A: ``\hl{You are correct that the question mentions a photograph of the distal esophagus. However, the answer choice C, ``Excessive transient lower esophageal relaxations," is the most likely cause of this patient's symptoms based on the clinical history provided in the question, regardless of the findings on the photograph. The patient's 10-year history of heartburn that occurs after eating late at night or consuming spicy food, and the temporary relief of symptoms with antacids, are consistent with a diagnosis of gastroesophageal reflux disease (GERD). The most common cause of GERD is excessive transient lower esophageal relaxations (TLESRs), which is why answer choice C is the most likely cause of this patient's symptoms. The photograph of the distal esophagus would be useful for confirming the diagnosis of GERD and assessing for any complications, but it is not necessary for selecting the most likely cause of the patient's symptoms based on the clinical history provided in the question."}

}
\end{AIbox}
\caption{{\DV} is asked to answer and then explain how it (correctly) solved a question which involves a photograph, without seeing the photo.}
\label{fig:media-interactive}
\end{figure}

\subsection{Influence of Alignment and Safety Tuning}
\label{sec:tuning}

The results above demonstrate the capabilities of the publicly released version of GPT-4. As described by \cite{openai2023gpt4}, GPT-4 was developed in two major phases. The pretraining phase focused on maximizing the ability of the model to predict the next token in a document using a variety of data, both publicly available and privately licensed. In a subsequent post-training phase, the model was fine-tuned using RLHF \cite{christiano2017deep} to enhance its ability to follow instructions and its propensity for producing fair and safe outputs \cite{ouyang2022training}. Several stages of refinement were aimed at minimizing undesirable behaviors such as gender bias, toxic language and stereotyping, dangerous recommendations, and harmful manipulation. OpenAI's experiments indicated that the RLHF-centric refinement did not adversely affect the model's capabilities on the exams they tested \cite{openai2023gpt4}.

\begin{table}[H]
\centering
\caption{Performance comparison of the publicly released GPT-4 model with GPT-4-base.} 
\begin{threeparttable}
        \begin{tabular}{llcccc}
        \toprule 
        \multirow{2}{*}{{\parbox{4cm}{\centering Dataset}}} & \multirow{2}{*}{Component} & \multicolumn{1}{c}{GPT-4-base} & \multicolumn{1}{c}{GPT-4-base} & \multicolumn{1}{c}{{\DV}} & \multicolumn{1}{c}{{\DV}} \\ 
        & & \multicolumn{1}{c}{(5 shot)} & \multicolumn{1}{c}{(zero shot)} & \multicolumn{1}{c}{(5 shot)} & \multicolumn{1}{c}{(zero shot)} \\ 
        \midrule 
        \multirow{3}{*}{\parbox{4cm}{\centering USMLE\\Self Assessment}} & Step 1 & \textbf{86.72} & 85.38 & 85.21 & 83.46 \\
        & Step 2 & \textbf{91.50} & 90.62 & 89.50 & 84.75 \\
        & Step 3 & \textbf{85.23} & \textbf{85.23} & 83.52 & 81.25 \\
        \midrule
        \multirow{3}{*}{\parbox{4cm}{\centering USMLE\\Sample Exam}} & Step 1 & \textbf{85.71} & 84.87 & 85.71 & 80.67 \\
        & Step 2 & 85.00 & \textbf{86.67} & 83.33 & 81.67 \\
        & Step 3 & 92.70 & \textbf{93.43} & 90.71 & 89.78 \\
        \bottomrule
    \end{tabular}
\end{threeparttable}
\label{tab:base-model}
\end{table}

We gained access to the base model, which we refer to as GPT-4-base, to study potential differences in performance attributable to the alignment process. The results of this evaluation are presented in Tables \ref{tab:base-model} and \ref{tab:multimedQA-scores}. While GPT-4-base and the release model both exhibit consistently strong performance across all 14 experimental datasets under study, we observe a notable increase of 3-5\% when using the base versus the release model.

The experiments suggest that orienting the base model toward safety and instruction-following can influence performance on medical benchmarks. The observed diminishment of raw performance accompanying the alignment process frames directions for future research. Refinements to the fine-tuning procedures employed to shape GPT-4-base into the publicly released GPT-4 may be able to better navigate the tradeoff between safety and accuracy. Additionally, alternative fine-tuning techniques, such as incorporating expert domain knowledge or leveraging specialized medical datasets, may lead to further improvements in model performance without sacrificing safety and instruction-following capabilities. 

\section{Medical Challenge Benchmarks}
\label{sec:med-benchmarks}

We present benchmarks for four multiple-choice datasets 
 from MultiMedQA \cite{singhal2022large}. The benchmarks include MedQA, PubMedQA, MedMCQA, and medical components of MMLU. MultiMedQA contains three more datasets which are not tested here. The untested datasets are LiveQA, MedicationQA, and HealthSearchQA; all have long answer formats that require extensive expert analysis to determine answer quality. Additionally, HealthSearchQA does not appear to be publicly available.

\subsection{Results}


\begin{table}[H]
\begin{center}
\begin{threeparttable}
\centering
\caption{Performance of different models on multiple choice components of MultiMedQA  \cite{singhal2022large}. {\DV} outperforms GPT-3.5 and Flan-PaLM 540B on every dataset except PubMedQA. {\DV} and GPT-3.5 were prompted with zero-shot direct prompts.}

    \begin{tabular}{lcccc}
        \toprule 
        \multirow{2}{*}{Dataset}  & \multicolumn{1}{c}{GPT-4-base} & \multicolumn{1}{c}{\DV} & \multicolumn{1}{c}{GPT-3.5} & \multicolumn{1}{c}{Flan-PaLM 540B\tnote{*}} \\ 
        & \multicolumn{1}{c}{5 shot / 0 shot} & \multicolumn{1}{c}{5 shot / 0 shot} & \multicolumn{1}{c}{5 shot / 0 shot} & \multicolumn{1}{c}{few shot}\\ 
        \midrule 
        \textbf{MedQA} & & & & \\
        Mainland China & \textbf{78.63} / 74.34 & 75.31 / 71.07 & 44.89 / 40.31 & --\\
        Taiwan & \textbf{87.47} / 85.14 & 84.57 / 82.17 & 53.72 / 50.60 & -- \\
        US (5-option) & \textbf{82.25} / 81.38 & 78.63 / 74.71 & 47.05 / 44.62 & --  \\
        US (4-option) & \textbf{86.10} / 84.45 & 81.38 / 78.87 & 53.57 / 50.82 & 60.3\tnote{**}  \\
        \midrule 
        \textbf{PubMedQA} & & & & \\
        Reasoning Required & 77.40 / \textbf{80.40} & 74.40 / 75.20 & 60.20 / 71.60 & 79.0 \\
        \midrule 
        \textbf{MedMCQA} & & & & \\
        Dev & \textbf{73.66} / 73.42 & 72.36 / 69.52 & 51.02 / 50.08 & 56.5 \\
        \midrule 
        \textbf{MMLU} & & & & \\ 
        Clinical Knowledge & \textbf{88.68} / 86.79 & 86.42 / 86.04 & 68.68 / 69.81 & 77.0 \\
        Medical Genetics & \textbf{97.00} / 94.00 & 92.00 / 91.00 & 68.00 / 70.00 & 70.0 \\
        Anatomy & 82.96 / \textbf{85.19} & 80.00 / 80.00 & 60.74 / 56.30 & 65.2 \\
        Professional Medicine & 92.65 / \textbf{93.75} & \textbf{93.75} / 93.01 & 69.85 / 70.22 & 83.8 \\
        College Biology & \textbf{97.22} / 95.83 & 93.75 / 95.14 & 72.92 / 72.22 & 87.5 \\ 
        College Medicine & \textbf{80.92} / 80.35 & 76.30 / 76.88 & 63.58 / 61.27 & 69.9 \\
        \bottomrule 
    \end{tabular}
    \begin{tablenotes}
        \item[*] Sourced directly from \cite{singhal2022large}. We use Flan-PaLM 540B few-shot results as the most directly comparable setting to our experimental setup. The number of few shot prompts used by Flan-PaLM 540B varies per dataset (between 3 and 5).
        \item[**] We note that \cite{singhal2022large} reports a preliminary performance of 67.2\% here with Med-PaLM, a prompt-tuned variant of Flan-PaLM 540B, using an ensemble of chain-of-thought, few-shot prompts.
    \end{tablenotes}

\label{tab:multimedQA-scores} 
\end{threeparttable} \\
\end{center}
\end{table}

For the MedQA and MMLU datasets, we report stratified performance metrics across different sub-components of the benchmarks. The MedQA benchmark also includes examination questions from mainland China and Taiwan, and covers three languages: English, simplified Chinese, and traditional Chinese. The English/United States version of the dataset contains two variants: a standard version with 5 multiple choice answers, and a simplified version with only 4 options. We report on both variants across all models considered. Similar to prior observations \cite{openai2023gpt4}, we find that {\DV} continues to perform well on difficult questions presented in non-English languages (Table \ref{tab:multimedQA-scores}).

In addition to the above results, \cite{lievin2022can} tested InstructGPT and Codex (\texttt{text-davinci-002} and \texttt{code-davinci-002} in the OpenAI API, respectively) with a large variety of prompts on MedQA, PubMedQA, and MedMCQA. When using zero-shot direct prompts, InstructGPT scored 46.0 on MedQA, 73.2 on PubMedQA, and 44.0 on MedMCQA. The best results from \cite{lievin2022can} come from testing Codex with an ensemble of 100 chain-of-thought samples, in which Codex scores 60.2 on the USMLE component of MedQA, 59.7 on the dev set of MedMCQA, and 78.2 on PubMedQA. In contrast, {\DV} shows a large boost in performance on MedQA and MedMCQA with a much simpler zero-shot prompt, continuing the narrative that the effort needed to obtain great performance drops dramatically with each model generation.

\section{Calibration}

We now focus on {\DV}'s calibration, a measure of the agreement between the predicted probabilities of each answer's correctness and the true frequencies of the outcomes. Calibration of the likelihood of the correctness of answers, or any assertions generated by an LLM, is critical for applications in high-stakes domains like medicine.  A well-calibrated model can provide trustworthy and interpretable probabilities that reflect the confidence of the model. Conversely, a poorly calibrated model can mislead users with overconfident or underconfident predictions, which can have harmful consequences.  Thus, appropriate characterizations of uncertainty about the veracity of generated content is important when providing generations, such as diagnoses and therapy plans, to healthcare professionals and other consumers of the information. For example, the probability that a treatment will be successful can be used in an expected-value calculation weighing the risks and benefits of a course of therapy. Looking to future applications of LLMs in medicine and other high-stakes areas, well-calibrated probabilities of generated content would enable contributions of LLM output to expected-utility decision making. We note that good calibration is not the same as high predictive accuracy. Predictive models can be accurate, yet poorly calibrated \cite{niculescu2005predicting}.

\begin{figure}[H]
\centering
\includegraphics[width=\textwidth]{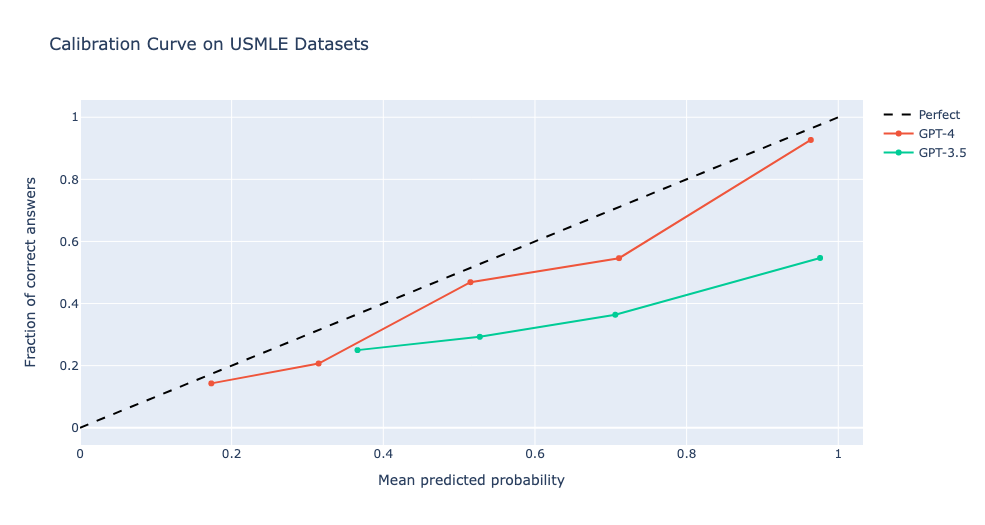}
\caption{Calibration comparison of {\DV} and GPT-3.5 on the USMLE Self-Assessment and Sample Exam.}
\label{fig:calibration-curve}
\end{figure}

A common method for measuring calibration is the aptly-named calibration plot. These plots bin predictions by their estimated probabilities, and measure how close the average probability in each bin is to the true positivity rate.  We adapt our multiple-choice question-answering setting to this framework by using the probability of the selected answer choice for each question. In Figure \ref{fig:calibration-curve}, we compare {\DV}'s calibration to that of GPT-3.5 on both official USMLE datasets. We find that {\DV} exhibits significantly better calibration than its predecessor on this type of data. For example, datapoints that {\DV} assigns an average probability of 0.96 tend to be correct 93\% of the time. In contrast, datapoints where GPT-3.5 assigns a similar probability are only correct 55\% of the time. 

We are only able to conduct this experiment in the multiple-choice question-answering setting where we have the model score each option. Measuring calibration and probabilities of long-form generation from generative models is an open area of research. However, our results on the multiple-choice problems suggest that the probability calibration of language models may increase with scale. 

\section{Directions and Limitations}

Our paper demonstrates the impressive potential of {\DV} to answer multiple choice questions on the USMLE, a medical licensing examination. We now review limitations and potential extensions of our studies.

\subsection{Richer Prompting Strategies}

A key goal of this paper is to establish baseline model performance metrics---that is, to determine how well {\DV} can perform {\em without requiring any specialized knowledge of LLMs and prompting}. Historically, more sophisticated prompting methods like chain of thought prompting \cite{wei2022chain, kojima2022large}, ensemble approaches like self-consistency prompting \cite{wang2022self}, or giving models access to information retrieval tools \cite{nakano2021webgpt} have all proven to significantly improve performance.
Furthermore, it is possible that new prompting patterns that work optimally for a new model like {\DV} have yet to be discovered. It is very likely that careful exploration of prompting techniques and fine tuning can achieve significantly higher performance numbers. As solely maximizing benchmark performance is not the goal of this paper, we largely leave those explorations to future work. We share the results of two preliminary experiments we did below on chain-of-thought prompting and expert curation of few shot examples.

\paragraph{Chain of Thought.} Following the work of \cite{kojima2022large} and \cite{wei2022chain}, we experiment with a two-stage chain-of-thought prompt on the USMLE sample exam. We first ask the model to ``think step by step" and lay out its reasoning. In a subsequent prompt, we then feed the entirety of the previous generation to {\DV} and ask for a final prediction. An illustrative example of the template is shown in Figure \ref{fig:cot-prompt-template}. As corroborated by other recent studies, we observe that basic chain-of-thought prompting does not yield performance benefits for {\DV} on medical questions. However, as shown in \cite{lievin2022can}, performance can vary significantly based on the specific chain-of-thought prompt used, and it is possible that a different prompt structure may yield stronger performance in future work.

\begin{figure}[H]
\begin{AIbox}{Chain-of-Thought Prompt template for multiple choice questions}
Question: \verb|{{question}}|

Choices:
\verb|{{answer_choices}}|
 \\
Let's think step by step. \\
\hl{\textit{initial model generation}} \\
Therefore, among A through \verb|{{last_choice}}|, the answer is 
\end{AIbox}
\caption{Two-stage chain-of-thought prompt template.}
\label{fig:cot-prompt-template}
\end{figure}

\paragraph{Few-shot example curation.} The authors of \cite{singhal2022large} worked with a panel of qualified clinicians to curate the best demonstration exemplars for the few-shot prompts, with custom prompts being designed for each dataset. We conducted light experiments comparing the exact curated few-shot exemplars sourced by \cite{singhal2022large} with our baseline random exemplar selection strategy on {\DV} and GPT-3.5. The performance difference between the two modes was negligible across the datasets we were able to test on. This finding suggests that expert exemplar curation may not be necessary to achieve strong performance in the latest generation of LLMs. We leave deeper investigations of this phenomena to future work.

\begin{table}[H]
\centering
\begin{threeparttable}
    \begin{tabular}{lcc}
        \toprule 
        \multirow{2}{*}{Dataset} & \multirow{2}{*}{\parbox{4cm}{\centering {\DV}\\(Random Exemplars)}} & \multirow{2}{*}{\parbox{4cm}{\centering {\DV}\\(Curated Exemplars)}} \\ 
        & & \\ 
        \midrule 
        MedQA US 5-option & \textbf{78.63} & 78.24 \\ 
        MedQA US 4-option & 81.38 & \textbf{82.33} \\
        MedMCQA & \textbf{72.36} & 71.36 \\
        PubMedQA & \textbf{74.40} & 74.00 \\
        \bottomrule 
    \end{tabular}
\end{threeparttable}
\label{tab:curated-examples} 
\caption{Random few-shot exemplar selection vs. expert curation.}
\end{table}

\subsection{Memorization}
\label{sec:memorization}

{\DV}'s strong performance on benchmark datasets raises the possibility that the system is leveraging \textit{memorization} or \textit{leakage} effects, which can arise when benchmark data is included in a model's training set. Given that LLMs are trained on internet-scale datasets, benchmark data may inadvertently appear in the model's training set. As the details of the training data for {\DV} are not public, other methods can be used to probe for memorization. We devised a heuristic algorithm to help identify potential signs of leakage through black-box evaluation of the model. With this approach, we prompt a model to generate a long set of near-exact matches to a given data sample and take similarity of the generation to the initial data as evidence of memorization. The method, which we refer to as memorization effects Levenshtein detector (MELD), can provide evidence that specific data is likely to be part of a model’s training set. Details of the procedure are provided in Appendix \ref{sec:memorzation-algorithm}.

We note that MELD has high precision but unknown recall. That is, if MELD detects a potential match, it is likely to be in the training set and memorized by the model. However, if our algorithm does not detect a match, it does not necessarily mean that the data was excluded from the training set. 

MELD is unable to find evidence of training data memorization in the official USMLE datasets we tested. Conversely, when we use MELD to test other popular datasets like SQuAD 2.0 and the Newsgroup Sentiment Analysis datasets, we are able to find strong evidence of existence in the training datasets. For example, {\DV} is able to regenerate questions from SQuAD 2.0 with 99\% overlap 17\% of the time, while it is completely unable to regenerate samples with even 50\% overlap on either of the USMLE datasets. We stress that this does not mean {\DV} has not seen this data before, only that we are unable to find evidence of it through our blackbox testing method.  

Given the findings obtained via the MELD procedure, and our sourcing of USMLE examination materials that are held behind an NMBE paywall, it is unlikely that official content in our examinations were in {\DV}'s training data. We further note that, even if contamination is present, {\DV}'s performance on USMLE examinations may not be significantly boosted. We note that OpenAI found that some contamination was prevalent across a large suite of publicly available benchmarks, but that the model did not perform differently on contaminated and uncontaminated data samples for the problems that were studied \cite{openai2023gpt4}.

\subsection{Focus on multiple choice} The benchmarking portions of this paper primarily focus on evaluating multiple choice exam questions, which constitute the majority but not entirety of the USMLE examinations. Specifically, Step 3 of the USMLE also includes 13 computer-based case simulations (CCS) that require candidates to manage patient care in a dynamic and interactive setting. While we qualitatively assess a hypothetical case simulation in Section \ref{sec:case} below, quantitative metrics on interactive challenges were not considered in the benchmarks. Furthermore, while we use a mixture of official and unofficial sources of multiple choice questions to test {\DV}, we do not have access to the actual USMLE questions used in recent exams or their scoring criteria. Therefore, the metrics we report may not be indicative of the true performance of {\DV} on a live USMLE exam.

\section{Beyond Correct Answers: Probing Capabilities}
\label{sec:case}

To move beyond statistical measures on exams and other benchmarks on medical challenge problems, we can qualitatively explore several capabilities of {\DV} by extending the challenge problems into interactive sessions. Beyond providing insights about the power of model, such extensions demonstrate directions with new forms of educational experiences and clinical applications that LLMs could enable.

We now share an illustrative case study of an interactive session nucleated by a single test question that demonstrates {\DV}'s capabilities for education. We initially provide {\DV} with a prompt and then simulate a dialog between the model and a medical student. On top of performing well on multiple-choice exam questions, {\DV} is able to provide rich explanations to the student about their errors. The model also exhibits the ability to hypothesize why the student may have made a mistake, and even conducts counterfactual analyses, delivering variants of the cinical problem with minimal rewrites to help simulate alternative scenarios. We note that the veracity of information generated in such interactions, as well as in real-world applications, cannot be trusted without review and confirmation by experts. We return to concerns with accuracy in Section \ref{sec:realworld}.

\subsection{Explanation}

\begin{figure}[H]
\begin{AIbox}{\DV: Initial prompt}
{\bf Prompt:} \\
{\scriptsize
A medical student is considering a challenging clinical situation: 

A 28-year-old man comes to the office with an episode of blood in his urine. He lived in sub-Saharan Africa until he came to the USA 6 months ago for graduate school. Temperature is 37.0°C, pulse is 60/min, respirations are 16/min, and blood pressure is 110/84 mm Hg. \\

\begin{tabular}{p{0.3\textwidth} p{0.3\textwidth} p{0.3\textwidth}} 
{\bf General} & {\bf Serum} & {\bf Urine} \\
Hemoglobin 12.3 g/dL & Urea nitrogen 75 mg/dL & Blood 3+ \\
Hematocrit 37\% & Creatinine 3.8 mg/dL & RBC 200/hpf \\
Leukocyte count 13,400/mm3 & & WBC 100/hpf \\
Segmented neutrophils 65\% & & RBC casts absent \\
Bands 5\% & & WBC casts absent \\
Eosinophils 5\% && \\ 
Lymphocytes 22\% && \\
Monocytes 3\% && \\
\end{tabular} \\

She is considering the most likely reason for these findings and is considering the following possibilities: 

(A) Exposure to a chemical toxin \\
(B) Interstitial cystitis \\
(C) Malacoplakia \\
(D) Schistosomiasis \\ 
(E) Vesicoureteral reflux \\  
(F) Acute Renal Failure \\
}
--- \\
{\scriptsize
\hl{The patient's history and lab results can help guide the medical student in determining the most likely cause of the clinical findings. \\
The patient is a 28-year-old man who lived in sub-Saharan Africa until 6 months ago, which is an important piece of information to consider. He presents with an episode of blood in his urine (hematuria). Lab results show elevated urea nitrogen and creatinine, indicating impaired kidney function. Additionally, there is an increased leukocyte count with a normal distribution of white blood cells. \\
Considering the patient's history and lab results, the most likely cause of these findings is:
(D) Schistosomiasis: This is a parasitic infection caused by Schistosoma species. The patient's history of living in sub-Saharan Africa is significant, as this region has a high prevalence of the disease. Schistosomiasis can cause hematuria and kidney dysfunction, which aligns with the patient's lab results. \\
 \\
The other possibilities are less likely due to various reasons: \\
(A) Exposure to a chemical toxin: While certain toxins can cause kidney dysfunction, there is no specific indication that the patient has been exposed to a toxin. Additionally, the patient's history and clinical presentation are more suggestive of schistosomiasis. \\
(B) Interstitial cystitis: This is a chronic condition characterized by bladder pain and urinary symptoms. While it can cause hematuria, it typically does not result in elevated urea nitrogen and creatinine levels. \\
(C) Malacoplakia: This is a rare inflammatory condition characterized by the presence of soft, yellowish plaques in the urinary tract. It can cause hematuria but is unlikely given the patient's history and lab results. \\
(E) Vesicoureteral reflux: This condition occurs when urine flows backward from the bladder into the ureters and kidneys. While it can cause kidney dysfunction, it is typically seen in children and is less likely in this case. \\
(F) Acute kidney injury: This is a possibility, but the patient's history and clinical presentation are more suggestive of schistosomiasis. Additionally, acute kidney injury usually presents with a more rapid decline in kidney function. \\
\\ 
In summary, schistosomiasis (D) is the most likely cause of the patient's symptoms and lab findings given his history and clinical presentation.
} \\

}
\end{AIbox}
\caption{{\DV} accurately diagnosing a clinical situation.}
\label{fig:education}
\end{figure}

\begin{figure}[H]
\begin{AIbox}{\DV: Explanation and teaching}
\centerline{\textit{interaction continued}} 
\centerline{...}
{\scriptsize
\hl{In summary, schistosomiasis (D) is the most likely cause of the patient's symptoms and lab findings given his history and clinical presentation.
} \\

$\rightarrow$ The student selects malacoplakia. Please explain to the student your assessment of why this is not the correct answer. \\

\hl{I understand that the medical student has chosen malacoplakia as the answer. However, let me explain why malacoplakia is not the most likely cause of the patient's symptoms and lab findings, based on the provided information. \\
\\
Malacoplakia is a rare, chronic inflammatory condition that affects the urinary tract. It is characterized by the presence of soft, yellowish plaques consisting of granulomatous tissue, usually in the bladder but occasionally in other parts of the urinary tract. Malacoplakia is often associated with immunosuppression, long-term antibiotic use, and urinary tract infections caused by Escherichia coli or other gram-negative bacteria. \\
\\
While malacoplakia can cause hematuria (blood in the urine), several factors make it less likely to be the cause in this particular case: \\
\\
Rarity: Malacoplakia is a rare condition, while the patient's history of living in sub-Saharan Africa until 6 months ago points towards a more common cause of hematuria in that region, such as schistosomiasis. \\
\\ 
Lab results: The patient's lab results show elevated urea nitrogen and creatinine levels, which indicate impaired kidney function. Although malacoplakia can occasionally affect the kidneys, it is more commonly localized to the bladder and not usually associated with significant kidney dysfunction.\\
\\
Clinical presentation: The patient's clinical presentation (hematuria, elevated urea nitrogen, and creatinine) does not specifically point towards malacoplakia. There is no mention of other urinary symptoms, such as pain, frequent urination, or a history of recurrent urinary tract infections, which might be more suggestive of malacoplakia. \\
\\
Patient's history: The patient's history of living in sub-Saharan Africa makes schistosomiasis a more likely cause of his symptoms, as the disease is endemic in that region and can cause both hematuria and kidney dysfunction. \\
\\
In summary, while malacoplakia can cause hematuria, it is a rare condition, and the patient's history, clinical presentation, and lab results are more consistent with schistosomiasis. Therefore, schistosomiasis remains the most likely cause of the patient's symptoms and lab findings.

}
}
\end{AIbox}
\caption{{\DV} explaining a medical diagnosis to a student.}
\label{fig:education-explanation}
\end{figure}

\subsection{Metacognition}

\begin{figure}[H]
\begin{AIbox}{\DV: Understanding the student}

\centerline{\textit{interaction continued}}
\centerline{...}

{\scriptsize
$\rightarrow$ Why might the student have selected malacoplakia? What findings in the case may have brought that to mind?  Can you share reflections on the reasoning that the student did to come up with this answer? \\

\hl{While I cannot definitively determine the medical student's thought process, I can offer some insights into why they may have considered malacoplakia as a possibility in this case. Here are a few factors that could have contributed to their selection: \\
\\
Hematuria: The presence of blood in the urine is a common symptom shared by various urinary tract conditions, including malacoplakia. The student might have focused on this symptom and considered malacoplakia as a potential cause without taking into account other factors, such as the patient's history or the likelihood of the disease. \\
\\
Differential diagnosis: In the process of generating a differential diagnosis, medical students are often taught to consider a wide range of possibilities, including rare conditions. The student may have included malacoplakia in their list of potential diagnoses and, perhaps due to a lack of experience or familiarity with the condition, overestimated its likelihood in this particular case. \\
\\
Knowledge gaps: Medical students are constantly learning, and it's possible that the student may not have had a full understanding of malacoplakia or its typical presentation. As a result, they may have mistakenly believed that the patient's symptoms and lab findings were consistent with this rare condition. \\
\\
Focus on lab results: The student may have been drawn to the patient's abnormal lab results, such as the elevated urea nitrogen and creatinine levels, and thought that malacoplakia could be a cause of these findings. However, the overall clinical picture, including the patient's history, is more consistent with schistosomiasis.
In summary, the medical student may have selected malacoplakia due to a combination of factors, such as focusing on the symptom of hematuria, considering a wide range of diagnoses, knowledge gaps, or placing too much emphasis on lab results. It's important for medical students to continually refine their diagnostic skills and learn from each case they encounter. This process will help them better understand the importance of considering the whole clinical picture, including the patient's history and the likelihood of each potential diagnosis, to arrive at the most accurate conclusion.
 
} \\

$\rightarrow$ Please explain to the student with care and understanding what the student may have thought but what the more likely diagnosis is and why?  Please be encouraging and supportive. \\

\hl{Dear student, \\
\\
I understand that you have considered malacoplakia as a possible cause of the patient's symptoms. It's important to explore various diagnoses, as this can help you refine your clinical reasoning skills. However, based on the information provided, there is another diagnosis that seems more likely in this case. \\
\\
The patient's history of living in sub-Saharan Africa until six months ago is a crucial factor to consider. Given this background, schistosomiasis becomes a more probable cause for the symptoms and lab findings. Schistosomiasis is a parasitic infection caused by Schistosoma species, which are more prevalent in sub-Saharan Africa. The disease can cause hematuria and kidney dysfunction, which aligns with the patient's elevated urea nitrogen and creatinine levels. \\
\\
While malacoplakia is a possibility, it is a rare condition, and its typical presentation doesn't fully align with the patient's clinical picture. The patient's history, symptoms, and lab findings are more consistent with schistosomiasis. \\
\\
It's essential to remember that, as a medical student, you are still learning and refining your diagnostic skills. Don't be discouraged by any missteps, as they are a natural part of the learning process. Instead, use them as an opportunity to grow and gain a deeper understanding of various conditions and their presentations.\\
\\
Keep in mind the importance of considering the patient's history and other factors when working through a differential diagnosis. This will help you weigh the likelihood of each potential diagnosis and ultimately lead you to the most accurate conclusion.
Keep up the good work and continue to learn from each case you encounter. Your dedication and curiosity will undoubtedly contribute to your growth as a medical professional.

}
}
\end{AIbox}
\caption{{\DV} demonstrates theory-of-mind---like ability to understand the student and personalize an educational explanation.}
\label{fig:metacognition}
\end{figure}

\subsection{Challenge Problem to Interactive Session}

\begin{figure}[H]
\begin{AIbox}{\DV: Counterfactual rewrites}

\centerline{\textit{interaction continued}}
\centerline{...}

{\scriptsize
$\rightarrow$ Can you change the history and/or lab results with as minimal changes as possible to make acute kidney injury (F) the best answer? Please list the changes that you have made after you write out the case.  Finally, summarize why the changes make acute kidney injury the best answer with the updates. \\

\hl{Certainly, here's the revised case with minimal changes: \\
\\
A 28-year-old man comes to the office with an episode of blood in his urine. He lived in sub-Saharan Africa until he came to the USA 6 months ago for graduate school.
Temperature is 37.0$^\circ$C, pulse is 60/min, respirations are 16/min, and blood pressure is 110/84 mm Hg.
}

\rowcolors{1}{aigreen}{aigreen} 
\begin{tabular}{p{0.3\textwidth} p{0.3\textwidth} p{0.3\textwidth}} \\
{\bf General} & {\bf Serum} & {\bf Urine} \\
Hemoglobin 12.3 g/dL & Urea nitrogen 75 mg/dL & Blood 3+ \\
Hematocrit 37\% & Creatinine 3.8 mg/dL & RBC 200/hpf \\
Leukocyte count 13,400/mm3 & & WBC 100/hpf \\
Segmented neutrophils 65\% & & RBC casts absent \\
Bands 5\% & & WBC casts absent \\
Eosinophils 5\% && \\ 
Lymphocytes 22\% && \\
Monocytes 3\% && \\
\end{tabular} \\

\hl{Changes made: \\
\\
The patient recently started using a nonsteroidal anti-inflammatory drug (NSAID) for joint pain. \\
 \\
With the addition of recent NSAID use, acute kidney injury (AKI) becomes a more likely explanation for the patient's symptoms and lab findings. Here's why: \\
\\
NSAIDs can cause acute kidney injury by reducing blood flow to the kidneys, leading to impaired kidney function. The patient's elevated urea nitrogen (75 mg/dL) and creatinine (3.8 mg/dL) levels indicate kidney dysfunction, which can be consistent with NSAID-induced AKI. \\
\\
Hematuria (blood in the urine) can occur in AKI, and the patient's recent NSAID use increases the likelihood of this being the cause.
The patient's leukocyte count is elevated, but there is no specific indication of an infection, making AKI more likely than an infectious cause, such as schistosomiasis. \\
\\
In summary, with the addition of recent NSAID use, acute kidney injury becomes the most likely explanation for the patient's symptoms and lab findings. It's important to note that this change alone doesn't completely exclude other possibilities but shifts the focus towards AKI as the best answer given the new information.}

}
\end{AIbox}
\caption{{\DV} minimally rewrites the case to create a counterfactual scenario.}
\label{fig:interactive-challenge}
\end{figure}

\begin{figure}[H]
\begin{AIbox}{\DV: Counterfactual rewrites (continued)}

\centerline{\textit{interaction continued}}
\centerline{...}

{\scriptsize
$\rightarrow$ Can you shift the labs to make AKI even more likely? \\
\\
\hl{Of course, here's the revised case with changes to the lab results to make acute kidney injury (AKI) even more likely: \\
\\
A 28-year-old man comes to the office with an episode of blood in his urine. He lived in sub-Saharan Africa until he came to the USA 6 months ago for graduate school.
Temperature is 37.0$^\circ$C, pulse is 60/min, respirations are 16/min, and blood pressure is 110/84 mm Hg.
}

\rowcolors{1}{aigreen}{aigreen} 
\begin{tabular}{p{0.3\textwidth} p{0.3\textwidth} p{0.3\textwidth}} \\
{\bf General} & {\bf Serum} & {\bf Urine} \\
Hemoglobin 12.3 g/dL & Urea nitrogen 120 mg/dL & Blood 3+ \\
Hematocrit 37\% & Creatinine 6.5 mg/dL & RBC 200/hpf \\
Leukocyte count 9,800/mm3 & & WBC 100/hpf \\
Segmented neutrophils 60\% & & RBC casts absent \\
Bands 3\% & & WBC casts absent \\
Eosinophils 2\% && \\ 
Lymphocytes 30\% && \\
Monocytes 5\% && \\
\end{tabular} \\
\\
\\
\hl{With these changes to the lab results, acute kidney injury (AKI) becomes even more likely for the following reasons: \\
\\
The patient's urea nitrogen (120 mg/dL) and creatinine (6.5 mg/dL) levels are now significantly elevated, which indicates a more severe kidney dysfunction that is consistent with AKI. \\
\\
The leukocyte count is still slightly elevated but closer to the normal range, and the eosinophils percentage has decreased, making an infectious or parasitic cause, such as schistosomiasis, less likely. \\
\\
The recent NSAID use remains an important risk factor for AKI, as these medications can cause kidney injury by reducing blood flow to the kidneys. \\
\\
In summary, with the changes made to the patient's history and lab results, acute kidney injury becomes the most likely explanation for the patient's symptoms and lab findings. The significant elevation in urea nitrogen and creatinine levels, along with the recent NSAID use, further supports AKI as the primary diagnosis.
}

}
\end{AIbox}
\caption{Per user request, {\DV} provides a sophisticated reformulation of the question focused by altering lab result values. (Case study continues in Appendix \ref{sec:app-case}.)} 
\label{fig:interactive-challenge-p2}
\end{figure}

\section{From Benchmarks to Real World}
\label{sec:realworld}

There is a large gap between competency on proficiency examinations and other medical benchmarks and the successful fielding and use of {\DV} in clinical applications. {\DV} and its successors have the long-term potential to exert a transformative influence on the field of medicine. Creative and appropriate uses of well-calibrated output could enhance the quality and reduce the costs of patient care. The performance on benchmarks, in addition to qualitative probes of its clinical capabilities, highlights the potential to assist physicians and other healthcare professionals with numerous aspects of healthcare delivery. The demonstrated competencies, particularly on USMLE Step 3, suggest that {\DV} and its successors can make contributions to clinical reasoning and daily workflows of medical practice. Beyond uses in decision support, memory jogging, and administrative tasks, {\DV} and its successors may one day assist investigators with clinical and biomedical research.  

\paragraph{Risks of erroneous generations.} Great care must be taken with the introduction of various forms of automation in healthcare, including uses of machine learning \cite{wiens2019no, caruana2015intelligible}. A critical concern is the accuracy of machine recommendations and their influence on decision makers. Methods and metrics have been developed for characterizing the performance of systems created via traditional supervised machine learning. With these applications, characterizing  overall accuracy of recommendations, as well as context- and instance-specific rates of error \cite{barraza2019error, erroranalysis, raitrackermitigationsblog, nori2019interpretml}, are enabled by the closed-world of well-defined, highly-focused problem areas, such as detecting a specific type of cancer \cite{esteva2017dermatologist, mckinney2020}, or predicting specific outcomes like readmissions \cite{bayati2014}, infection \cite{wiens2016patient}, sepsis \cite{adams2022}, and in-hospital deterioration \cite{escobar2020automated}.  Unfortunately, similar characterizations of reliability and confidence are not yet available for the massive, open-world of generations that are output from foundation models in response to prompts. Difficulties with evaluation of the output of LLMs in supporting real-world decisions include the challenge of stability and robustness of recommendations and inferences generated in response to custom-tailored prompting in the wild. Generations are often highly sensitive to details of the wording of prompts. Stability of generations can also be sensitive to model revision that might be ongoing, including rebuilding or fine-tuning of models based on fresh data \cite{srivastava2020empirical}.

While large language models show promise with providing support in healthcare administration and delivery, caution is needed to mitigate potential negative influences of over-reliance on model-generated recommendations. Significant risks with uses of large language models include inaccurate recommendations about rankings (e.g., with differential diagnoses) and sequencing (e.g., information gathering, testing), as well as blatant factual errors, particularly with important omissions and with erroneous generations, often referred to as {\em hallucinations}. LLM hallucinations can be particularly difficult to detect given the high linguistic fluency of the models and the ability to interleave inaccurate and ungrounded assertions with accurate generations. Such hallucinations can include incorrect or misleading medical information which necessitates careful review and fact checking. Thus, extreme caution is required when using LLMs in high-stakes medical applications, where incorrect or incomplete information could have serious consequences for patient care.

Additional research is needed to address the veracity of model output. Directions include employing search and retrieval to help ground generations in the literature, doing checks of self-consistency, performing evaluation studies to characterize the overall statistics of accurate generations, conditioned on different contexts and usages, and refinements of methods that generate and harness accurate calibration signals. Trusted calibration information can be harnessed in numerous ways to enable practitioners to better understand and assess model outputs\cite{bansal2019updates, bansal2019beyond, wilder2020learning}. More generally, innovation with human-computer interaction will be valuable in the fielding of applications of LLMs in healthcare \cite{fogliato2022goes,amershi2019guidelines,principles1999, mozannar2022}. 

Healthcare providers relying on {\DV} and other models will need to adhere to the highest standards for verifying information generated by the model. Best practices for quality assurance must be developed and shared among medical professionals to ensure safe and effective use.  In the context of erroneous omissions and inclusions, healthcare professionals and other consumers of health-related content will need to be educated on the challenges with reliability and the need for ongoing vigilance. Education, awareness, and promoting guidelines for best practices may help to minimize safety challenges.

\paragraph{Risks of Bias.} Studies have revealed biases in the delivery of healthcare, noting disparities in care received by people experiencing marginalization \cite{Charethical18,hall2015implicit}. Such biases in healthcare delivery have been demonstrated to influence the systems and models that are developed to provide guidance to healthcare organizations and practitioners. Without study to address and mitigate bias in data and the systems constructed from that data, we risk fielding systems that propagate long-term disparities and inaccuracies \cite{Obermeyer2019}. Exploration of biases in healthcare and the potential reflection of these biases in AI systems come in the context of broader work on biases of AI systems. Several efforts have demonstrated that the output of machine-learned models can be unfair and harmful to specific groups of people, depending on backgrounds and demographics (e.g., \cite{howard2017addressing,buolamwini2018gender}). Progress has been made on technical methods for detecting and mitigating harmful biases in task-specific systems built via supervised machine learning (e.g., \cite{pmlr-v80-agarwal18a, kulshrestha2021fairlearn}). However, engineers and organizations continue to face conceptual challenges with defining measures of fairness \cite{jacobs2021measurement, barocas-hardt-narayanan}.

We have a poor understanding of the biases accrued and transmitted by large-scale language models, and how issues with fairness might arise for different types of healthcare-centric prompting and generations. In the absence of study, we must be wary of biases in both clinical practices and research \cite{agmon2022gender} with regard to race, socioeconomic background, gender, and other factors, which are laced throughout the corpora used to train large-scale language models. Research is needed to understand the fairness of healthcare-centric recommendations generated by LLMs.

\paragraph{Influences on workflows, tasks, and specialties.} The competencies on the USMLE examinations and other medical workloads suggest that {\DV}, properly harnessed with appropriate expert oversight, can contribute to enabling precision clinical medicine. {\DV} and its successors could be leveraged to provide healthcare practitioners with analytics, reminders, and decision support, including assistance with the formulation and revision of differential diagnoses from patient history, physical findings and lab results, identification of relevant tests and their sequencing, and constructing therapy plans. With effective management of errors, LLMs could help with memory jogging, alerting, and screening. As an example, memory jogging might help to mitigate persistent challenges with adverse outcomes, including those attributable to preventable human errors \cite{bates2023safety,grimm2022adverse} and diagnostic and therapeutic delays \cite{newman2022diagnostic}. 

In the longer-term, {\DV} and its descendants might be harnessed to shift the distribution of tasks associated with daily flows of work for physicians and other healthcare practitioners in ways that reduce the burden of programmatic, logistical, and administrative tasks.
Reduction in the drudgery of writing reports and performing other administrative tasks would permit healthcare providers to spend more time on the uniquely human elements of the profession, such as patient engagement and coordinating and collaborating with healthcare colleagues. The technology could also enable more time for physicians to learn, reflect, and engage in continuing medical education to become the best at what they are interested in doing. In addition, LLMs could be harnessed to provide information, communication, screening, and decision support in under-served regions. The models could help to raise the competency of physicians' assistants and help with triage and communication with remote experts.

\paragraph{Social and societal issues.} On a broader social and societal front, the capabilities demonstrated by {\DV} can influence decisions about the choice to pursue a medical career, the choice of residency and ultimate speciality, and the sense of uniqueness of human contributions for today's healthcare practitioners \cite{topol2019deep}. 
 AI's accelerating performance on competency exams and other medical challenge problems may contribute to the perception that the technology's inroads in medicine will eventually devalue human intellect. Practitioners may be concerned about significant shift in the way medical specialties are practiced or valued. At earlier steps in the chain of training and commitment, AI's growing competence could influence career choices in medicine, shifting perceptions of which tasks rely on genuine human intellect. This may change decisions about medicine as a career path overall and, for those already in medical training programs, their choice of speciality. Perhaps as early foreshadowing, a recent study found that medical students' choice of radiology as a career is significantly reduced by their perception of the growing role of AI in radiology \cite{Reeder2022}.

\paragraph{Implications for the future.} The leap in performance on medical challenge problems with the move from GPT 3.5 to {\DV} suggests that we can achieve impressive gains on intensive real-world challenges with scale--and that we will likely continue to see advances with larger models for handling complex, real-world problems.  The rate of progress of LLMs has implications beyond the medical profession. As observed in \cite{susskind2018future}, large swaths of modern society are predicated on a ``grand bargain" in which professional classes invest years or even decades in technical education and training and are, in turn, afforded certain benefits by citizens and governments, including the exclusive right to practice in their field, social prestige, and above-average compensation. Technical disruption of this social contract can have implications not only for the medical field but for numerous other knowledge-intensive professions including law, banking, engineering, accounting, and others. 
 
\section{Conclusion}

 We presented a comparative evaluation of {\DV}, GPT-3.5 and Flan-PaLM 540B on medical competency examinations and benchmark datasets. We explored zero-shot performance as a baseline. We first focused on answers to questions that are representative of those included in USMLE Step 1, Step 2, and Step 3 exams, certification tests given to medical students and residents in the U.S. We studied performance on questions relying solely on text versus questions referring to visual media. We found that {\DV} significantly outperforms GPT-3.5 and Flan-PaLM 540B. Next, we demonstrated that {\DV} significantly surpasses GPT-3.5's performance on the MultiMedQA dataset. The model also outperformed Flan-PaLM 540B on all but one dataset. Our work also explored the calibration of the model's output probabilities, highlighting the importance of calibration for medical applications. We provided sample output that demonstrates {\DV}'s capacity for reasoning about the concepts tested in USMLE challenge problems, including explanation, counterfactual reasoning, differential diagnosis, and testing strategies. Finally, we touched on broader implications of applications of {\DV} in medicine. 
 We maintain that {\DV}'s exceptional performance on benchmarks serves as an indicator of its potential for being harnessed in medical education and for aiding healthcare professionals with numerous aspects of healthcare delivery. Nonetheless, considering the possibility of errors and the challenges in assessing performance in real-world scenarios, it is vital to practice prudence, seek to develop and evaluate appropriate uses, and pursue technical innovations to optimize advantages and mitigate risks associated with applications.
 
\section*{Acknowledgments}

We thank Miles Brundage, Pamela Mishkin, and Jack Rae for their feedback and assistance on this effort. We are grateful for insights shared in earlier discussions by Rich Caruana, Scott Lundberg, Matthew Lungren, Marco Tulio Ribeiro, and Nigam Shah.

\bibliographystyle{alpha}
\bibliography{mainbib}

\appendix

\section*{Appendix}

\section{Dataset descriptions}
\label{sec:dataset-descriptions}

\begin{itemize}
    \item \textbf{USMLE Self Assessments}: This dataset includes materials purchased from the NBME Self-Assessment resources available at \url{https://www.nbme.org/examinees/self-assessments}. The dataset contains 2173 questions in total, with 1197 questions in Step 1, 800 questions in Step 2, and 176 questions in Step 3. Step 1 exam forms include Form 25, Form 26, Form 27, Form 28, Form 29, and Form 30. Step 2 exam forms include Form 9, Form 10, Form 11, and Form 12. Step 3 exam forms include Form 1, Form 2, Form 3, Form 4, and Form 5.
    \item \textbf{USMLE Sample Exam}: Sample exam materials were sourced from USMLE practice materials at \url{https://www.usmle.org/prepare-your-exam}. Exam materials are contained in the following PDFs. Step 1: \url{https://www.usmle.org/sites/default/files/2021-10/Step_1_Sample_Items.pdf}. Step 2: \url{https://www.usmle.org/sites/default/files/2021-10/Step2_CK_Sample_Questions.pdf}. Step 3: \url{https://www.usmle.org/sites/default/files/2021-10/Step3_Sample_Items.pdf}. This dataset is the same as the one used by Kung et al. to test ChatGPT \cite{kung2023performance}, but in our tests, the questions containing media were not removed from the test set.
    \item \textbf{MedQA}: MedQA contains English, Simplified Chinese, and Traditional Chinese multiple-choice questions collected from Medical Licensing Examinations of the United States, Mainland China, and Taiwan \cite{jin2021disease}. There are 5 options per question in the United States and Mainland China segments, with the following number of questions in the test split: 1273 US, 3426 Mainland. The Taiwan segment has 1413 questions with 4 options per question. The US test split also has a simplified 4 option version with the same 1273 questions, where one of the incorrect options is removed. 
    \item \textbf{PubMedQA}: PubMedQA contains tests requiring a model to answer yes, no, or maybe to biomedical research questions when given context provided from PubMed abstracts \cite{jin2019pubmedqa}. There are two settings for PubMedQA tests called reasoning-required and reasoning-free. In the reasoning-free setting, a long-form answer that contains explanations of the abstracts is provided. We report results for the reasoning-required setting, in which the model is only given context from abstracts to use when answering the question. The yes, no, or maybe choices are formulated as a multiple choice question, where A is yes, B is no, and C is maybe, matching the testing method done by Li{\'e}vin et al. \cite{lievin2022can}
    \item \textbf{MedMCQA}: MedMCQA is a dataset of multiple choice questions sourced from mock exams and published past exams of two Indian medical school entrance exams called AIIMS and NEET-PG \cite{pal2022medmcqa}. The dev set contains 4183 questions, and the test set contains 6150 questions. Each question has 4 choices.
    \item \textbf{MMLU}: MMLU is a large multiple choice question dataset spanning a wide variety of topics \cite{hendrycks2020measuring}. MMLU has 57 subjects in total. Results of medicine related subjects are reported in our benchmarks.
\end{itemize}

\section{Memorization analysis}
\label{sec:memorzation-algorithm}

The memorization effects Levenshtein detector (MELD) method works as follows: for each content datapoint in a dataset, we partition the content in half. We pass in the first half as context and ask the model to generate up to the length of the second half. We then measure how similar the generated content is to the held-out second half using Levenshtein distance ratio, which is defined as one minus the ratio of Levenshtein distance to the maximum possible distance. We define a near-exact match as having a Levenshtein distance ratio of at least 0.95, meaning that less than 5\% of the characters are different between the generated and held out content. We measure the rate of near exact matches by computing the percentage of datapoints that have a near-exact match in each dataset. While each individual sample may not be regenerated exactly due to stochastics in the training process, the observed rate across the entire dataset can provide strong empirical evidence that the model has seen the data before.

\vspace{.3in}

\begin{algorithm}[H]
 \caption{Memorization effects Levenshtein detector (MELD)}
 \KwData{A generative language model $g$, a dataset $D$ of question-answer pairs, a tokenizer $T$, a threshold $Y$}
 \KwResult{A metric $Z$ that measures the percentage of completions that have more than $Y\%$ overlap with the ground truth}
 \vspace{0.1in}
 Initialize an empty list $L$\;
 \ForEach{$(q,a) \in D$}{
  Split $q$ into two halves: $q_1$ and $q_2$\;
  Tokenize $q_1$ and $q_2$ using $T$:  $t_1 = T(q_1)$ and $t_2 = T(q_2)$\;
  Set the model temperature to zero and pass $q_1$ as context\;
  Generate $k$ tokens from $g$ where $k = |t_2|$\;
  Calculate the Levenshtein distance-based ratio between $t_2$ and generated text $g$: 
    $$\ell = \frac{\text{int}(\text{round}(\frac{2.0 \times M}{|q|} \times 100))}{100}$$
    where $|q|$ is the total number of characters in both strings, and $M$ is the number of matches in the two strings.\;
  \If{$\ell > Y$}{
   Append $\ell$ to L\;
   }
 }
 Calculate the mean of L: Z = mean(L)\;
 Return Z\;
\end{algorithm}

\section{Additional prompt examples}
\label{sec:app-prompts}

\begin{figure}[H]
\begin{AIbox}{GPT-4 chat prompt template for multiple choice questions}
\begin{verbatim}
[
  { 
    role: "system",
    content: "You are a helpful assistant that answers multiple choice
              questions about medical knowledge."
  },
  {{few_shot_examples}}
  {
    role: "user",
    content: "{{context}}**Question:** {{question}} {{answer_choices}}"
  },
  {
    role: "assistant",
    content: "**Answer:**("
  }
]
\end{verbatim}
\end{AIbox}
\caption{Template used to generate prompts for GPT-4 on all multiple choice questions. Elements in double braces \{\{\}\} are replaced with question-specific values. Few shot examples are encoded as user and assistant chat messages.}
\label{fig:chat-prompt-template}
\end{figure}

\begin{figure}[H]
\begin{AIbox}{Sample question using chat prompt template}
\begin{verbatim}
[
  {
    role: "system",
    content: "You are a helpful assistant that answers multiple choice
              questions about medical knowledge."
  },
  {
    role: "user",
    content: "**Question:** A 72-year-old woman who has smoked 20
              cigarettes daily for the past 38 years begins using
              eyedrops for glaucoma. Three days later, she has a
              marked increase in shortness of breath while walking
              up a flight of stairs. Which of the following drugs is
              the most likely cause of the development of shortness
              of breath in this patient?
              (A) Apraclonidine
              (B) Epinephrine
              (C) Latanoprost
              (D) Timolol"
  },
  {
    role: "assistant",
    content: "**Answer**:("
  }
]
\end{verbatim}
\end{AIbox}
\caption{Instantiated example of Figure \ref{fig:prompt-template}.}
\label{fig:chat-prompt-example}
\end{figure}

\section{Interactive educational session (continued)}
\label{sec:app-case}

\begin{figure}[H]
\begin{AIbox}{\DV: Formulating differential diagnosis}

\centerline{\textit{interaction continued}}
\centerline{...}

{\scriptsize
$\rightarrow$ Going back to the original case, what is the differential diagnosis, considering the broader possibilities, including if necessary, but going beyond the multiple choice answers? Please rank by likelihood. \\
\\
\hl{
Considering the broader possibilities for the differential diagnosis of the original case, we can list the following conditions and rank them by likelihood: \\
\\
Schistosomiasis (most likely): The patient's history of living in sub-Saharan Africa, presence of hematuria, and kidney dysfunction make this the most likely cause.
Acute kidney injury (AKI): Elevated urea nitrogen and creatinine levels indicate kidney dysfunction, which could be due to AKI from various causes, such as infection, ischemia, or exposure to nephrotoxic agents. \\
\\
Glomerulonephritis: Inflammation of the glomeruli in the kidneys can cause hematuria and kidney dysfunction. The patient's lab findings could be consistent with this condition. \\
\\
Kidney stones: Although less likely, kidney stones can cause hematuria and kidney dysfunction in some cases. \\
\\
Exposure to a chemical toxin: Certain toxins can cause kidney dysfunction and hematuria, but there is no specific indication in the case that the patient has been exposed to a toxin. \\
\\
Interstitial cystitis: A chronic condition characterized by bladder pain and urinary symptoms. While it can cause hematuria, it typically does not result in elevated urea nitrogen and creatinine levels. \\
\\
Malacoplakia: A rare inflammatory condition characterized by the presence of soft, yellowish plaques in the urinary tract. It can cause hematuria but is unlikely given the patient's history and lab results. \\
\\
Vesicoureteral reflux: This condition occurs when urine flows backward from the bladder into the ureters and kidneys. While it can cause kidney dysfunction, it is typically seen in children and is less likely in this case. \\ 
\\
Tubulointerstitial nephritis: Inflammation of the renal tubules and interstitium can cause kidney dysfunction and sometimes hematuria. However, the patient's history and lab results do not strongly suggest this condition. \\
\\
In the original case, schistosomiasis remains the most likely cause of the patient's symptoms and lab findings, given his history and the overall clinical picture. Other possibilities like AKI, glomerulonephritis, and kidney stones should be considered and evaluated based on additional information and further diagnostic tests.

}
}
\end{AIbox}
\caption{Case study continued: When pressed further, {\DV} provides a more sophisticated rewrite focused on altering lab result values.} 
\label{fig:differential-diagnosis}
\end{figure}

\begin{figure}[H]
\begin{AIbox}{\DV: Information-gathering and testing}

\centerline{\textit{interaction continued}}
\centerline{...}

{\scriptsize
$\rightarrow$ Given the differential diagnosis, what additional information from the patient and lab tests would be most helpful to narrow down the possibilities--and how should tests be ideally sequenced? \\
\\
\hl{
To narrow down the differential diagnosis for the patient's symptoms and lab findings, additional information from the patient and further lab tests can be conducted. Here's a suggested sequence of tests and information gathering: \\
Detailed patient history: Start by obtaining a more detailed medical, travel, and social history, including any medication or supplement use, recent illnesses, exposures to infections, or possible contact with toxins. \\
\\
Urinalysis: A more comprehensive urinalysis, including urine protein levels, microscopic examination for red and white blood cell casts, and a urine culture to identify possible infections, can provide valuable information on the underlying cause of the patient's symptoms. \\
\\
Serology for schistosomiasis: Given the patient's history of living in sub-Saharan Africa, testing for Schistosoma antibodies can help confirm or exclude schistosomiasis as the cause of the patient's symptoms. \\
\\
Kidney ultrasound: Imaging of the kidneys can provide information about their size, shape, and possible structural abnormalities. It can also identify hydronephrosis, kidney stones, or an obstructive cause for AKI. \\
\\
Additional blood tests: Measure serum complement levels, antinuclear antibody (ANA), anti-neutrophil cytoplasmic antibody (ANCA), and anti-glomerular basement membrane antibody (anti-GBM) to assess for possible glomerulonephritis. \\
\\
Cystoscopy: If the previous tests do not lead to a clear diagnosis, a cystoscopy can be performed to visualize the bladder and assess for interstitial cystitis, malacoplakia, or any other abnormalities. \\
\\
Kidney biopsy: In cases where the diagnosis remains uncertain, a kidney biopsy can help identify the underlying cause of the patient's kidney dysfunction, such as glomerulonephritis or other renal pathologies. \\
\\
The ideal sequencing of tests depends on the individual case, availability of resources, and the clinical judgment of the healthcare provider. It is important to prioritize tests based on the most likely diagnoses given the patient's history and clinical presentation. In this case, a detailed patient history, urinalysis, serology for schistosomiasis, and kidney ultrasound are the initial tests that can provide valuable information to narrow down the differential diagnosis. Further tests, such as blood tests for autoimmune markers and kidney biopsy, can be performed if the initial tests do not provide a definitive diagnosis.
}

}
\end{AIbox}
\caption{{\DV} offers recommendations on information and tests that can narrow the differential diagnosis.} 
\label{fig:differential-diagnosis-followup}
\end{figure}

\begin{figure}[H]
\begin{AIbox}{\DV: Schistosomiasis and Patient Conversations}

\centerline{\textit{interaction continued}}
\centerline{...}

{\scriptsize
$\rightarrow$ What are ideal treatments for schistosomiasis? \\
\\
\hl{
Schistosomiasis is a parasitic infection caused by various species of flatworms called Schistosoma. The ideal treatment for schistosomiasis primarily involves the use of antiparasitic medications. The choice of medication and duration of treatment depend on the specific Schistosoma species causing the infection. \\
\\
Praziquantel: Praziquantel is the drug of choice for treating all major forms of schistosomiasis, including those caused by Schistosoma mansoni, Schistosoma haematobium, and Schistosoma japonicum. It works by paralyzing the parasites and making them susceptible to the host's immune system. Praziquantel is typically given as a single oral dose or split into multiple doses taken on the same day, depending on the species and severity of the infection. \\
\\
Oxamniquine: Oxamniquine is an alternative treatment for Schistosoma mansoni infections, particularly in areas where Praziquantel resistance has been reported. However, it is not effective against other species of Schistosoma. \\
\\
In addition to antiparasitic medications, supportive care may be necessary in some cases, depending on the severity of the infection and the presence of complications: \\
Analgesics or anti-inflammatory medications: These drugs can help manage pain and inflammation associated with the infection. \\
\\
Corticosteroids: In severe cases with complications such as central nervous system involvement or pulmonary hypertension, corticosteroids may be used to manage inflammation and reduce symptoms. \\
\\
Treatment of complications: Specific complications, such as anemia, secondary infections, or organ damage, may require additional medical management. \\
\\
Monitoring and follow-up: After treatment, patients should be monitored for resolution of symptoms and clearance of the infection. In some cases, repeat treatment with antiparasitic medications may be necessary if the infection persists or reinfection occurs. \\
\\
Preventive measures, such as access to clean water, sanitation improvements, and snail control, are essential to reduce the risk of schistosomiasis transmission in endemic areas. Mass drug administration with Praziquantel is also used as a public health strategy to control the disease in high-risk populations. 
} \\
 \\
$\rightarrow$ What might the physician treating the patient say in a caring way to the patient when revealing the likely diagnosis? \\
 \\
\hl{
Doctor: [Patient's name], thank you for providing us with all the information about your symptoms and medical history. Based on your recent experience of living in sub-Saharan Africa, your symptoms, and the lab results, it appears that you may have a parasitic infection called schistosomiasis. This condition can cause blood in your urine and affect your kidney function, which is consistent with your lab findings. \\
\\
I understand that receiving a diagnosis like this can be concerning, but I want to assure you that schistosomiasis is treatable. We will prescribe an antiparasitic medication called Praziquantel, which is very effective in treating this infection. You should start to see an improvement in your symptoms after completing the treatment. \\
\\
It's important to follow the treatment plan and attend follow-up appointments so we can monitor your progress and ensure the infection has been cleared. If you have any questions or concerns, please don't hesitate to ask. We are here to support you throughout your treatment and recovery. \\
\\
Remember, you're not alone in this journey, and we'll do everything we can to help you regain your health.
}

}
\end{AIbox}
\caption{Advice from {\DV} on engaging with a patient with the diagnosis of schistosomiasis.} 
\label{fig:schistosomiasis}
\end{figure}

\end{document}